\documentclass[lettersize,journal]{IEEEtran}
\usepackage{amsmath,amsfonts,amssymb}
\usepackage{algorithmicx}
\usepackage{algorithm}
\usepackage{algpseudocode}
\usepackage{array}
\usepackage[caption=false,font=footnotesize]{subfig}
\usepackage{textcomp}
\usepackage{stfloats}
\usepackage{url}
\usepackage{verbatim}
\usepackage{graphicx}
\usepackage{booktabs}
\usepackage{tabularx} 
\usepackage{makecell} 
\usepackage{multirow} 
\usepackage{enumitem}
\usepackage{xurl}  
\usepackage[hidelinks]{hyperref} 
\newcommand{\best}[1]{\textbf{#1}}
\newcommand{\second}[1]{\underline{#1}}
\usepackage{cite} 
\newcolumntype{Y}{>{\centering\arraybackslash}X}

\begin{document}
\title{Navigating Potholes: Learning Safe Motion Planning with Vehicle Dynamics}

\author{Xiang~Wang,
        Scott~Piersall,
        Zihang~Zou,
        Liqiang~Wang,
        and~Rongjie~Yu% <-this % stops a space
\thanks{Xiang Wang and Scott Piersall contributed equally to this work. \textit{(Corresponding author: Rongjie Yu.)}}%
\thanks{Xiang Wang and Rongjie Yu are with the College of Transportation, Tongji University, Shanghai 201804, China (e-mail: \textit{yurongjie@tongji.edu.cn}).}%
\thanks{Scott Piersall and Liqiang Wang are with the University of Central Florida, Orlando, FL, USA.}%
\thanks{Zihang Zou is with Optixway AI, USA.}%
}

\maketitle

\begin{abstract}

Potholes are one of the most common and critical road-surface hazards that challenge safe autonomous driving. Although autonomous vehicles (AVs) can detect potholes, they mainly respond by braking to reduce pothole impact, which may increase the risk of rear-end collisions. This braking-based strategy can also overlook feasible in-lane lateral maneuvers for safer pothole handling. To address this problem, we propose a novel motion planning framework that jointly uses lateral maneuvers and speed control for safer and more efficient pothole handling. The framework integrates a quarter-car vertical dynamics model and a physics-based tire-pothole impact model to estimate wheel loads, ride comfort, and structural safety limits. A Soft Actor-Critic (SAC) agent then learns decisions on lateral maneuvers and speed control that balance safety, comfort, and efficiency, while a curriculum learning strategy improves training under sparse impact feedback and diverse pothole geometries. Simulation results show that the proposed method improves pothole avoidance, structural safety, and rear-end safety over the baselines. In scenarios where lateral avoidance is feasible, it achieves a 92.5\% avoidance rate and reduces travel time by about 50\% compared with braking-based methods. In scenarios where avoidance is not feasible, it lowers the rear-end collision rate from 13.3\% to 1.7\% while achieving a 46.7\% structural safety rate. These results suggest that the proposed framework is promising for advanced motion planning with vehicle dynamics, facilitating the development, deployment, and safe operation of AVs.

\end{abstract}

\begin{IEEEkeywords}
Autonomous vehicles, motion planning, pothole, reinforcement learning, vehicle dynamics. 
\end{IEEEkeywords}

% ==========================================
% SECTION 1: INTRODUCTION
% ==========================================
\section{Introduction}

\IEEEPARstart{A}{utonomous} vehicles (AVs) are advancing towards extensive real-world deployment, evidenced by commercial robotaxi services worldwide~\cite{Waymo2025FleetScale,ReutersWaymo2026Funding,PRNewswireBaiduDriverlessPermits2022,PRNewswireBaiduBeijingPermit2023} and regulatory approvals for SAE Level~3 conditional automation in Japan, Germany, China, and the United States~\cite{Honda2020TypeDesignation,Mercedes2024DrivePilot95,CADMV2023DrivePilotPermit,ReutersChinaL3Approval2025}. AV systems primarily focus on interaction-centric objectives, ensuring safety and efficiency in traffic scenarios. However, they often overlook vehicle-dynamics-related challenges, such as those posed by off-nominal road surfaces, including low-adhesion conditions (e.g., rain or snow) and physically defective surfaces (e.g., potholes).

Potholes are prevalent across diverse road networks, affecting approximately 44 million U.S. drivers annually~\cite{AAAEastCentral2024PotholesWallet}. Pothole-related damage results in nearly \$3 billion in annual repair costs, with per-incident expenses reaching hundreds of dollars~\cite{AAAEastCentral2024PotholesWallet,AAAWCNY2023MassiveBump}. This extensive damage stems from the abrupt geometric discontinuity of potholes, causing severe tire-road edge collisions that generate high-magnitude transient loads~\cite{radu2023development,Zhang2020TireObstacle}. These forces can compromise wheel components and transmit substantial impulsive forces to the suspension structure, highlighting the urgent need for effective countermeasures.

Existing sensing technologies, such as cameras and LiDAR, can effectively identify potholes and extract relevant information like dimensions and depth~\cite{MercedesRoadSurfaceScan,JLRPotholeDetection,Yebes2021PotholeDetection,Abbasi2025LidarSegmentation,xiang2023multi}. Vehicle-to-cloud (V2X) sharing and integration with navigation systems further extend this awareness~\cite{Han2023CollaborativePerception,xiang2023multi}. Researchers have explored control-side countermeasures using predictive/active suspension to suppress body vibration and mitigate impact-induced excitations~\cite{MercedesRoadSurfaceScan,AudiA8PredictiveSuspension,NIO_SkyRide,BYDDiSus}. However, active suspension remains limited due to high costs, low market penetration, and actuator stroke/bandwidth constraints, which restrict its effectiveness under severe impacts.

With the evolution of steer-by-wire and brake-by-wire actuation, alongside advancements in automated driving, researchers have increasingly investigated vehicle motion planning and control for pothole handling~\cite{Du2022RoughSpeedDRL,Zhang2023BumpyITSC,Huang2021SharedControl}. Existing studies primarily focus on longitudinal speed regulation, active suspension control, and their integration. However, these approaches seldom exploit the potential of intra-lane lateral redundancy to actively bypass or straddle hazards, missing substantial dynamics-related performance benefits~\cite{Zhang2020TireObstacle}.

Deep Reinforcement Learning (DRL) offers a promising alternative to address these online optimization complexities by learning from environment interactions and learning coordinated handling strategies that balance conflicting objectives~\cite{Aradi2022SurveyDRLMotionPlanning, Du2022RoughSpeedDRL, Zhao2024ReachabilityGuidedRL,Behzadan2021AdversarialRL,wang2024survey}. However, deploying DRL in vehicle-dynamics-oriented tasks presents unique challenges due to strict physical failure boundaries, which can destabilize learning dynamics and trap policies in sub-optimal local minima~\cite{Aradi2022SurveyDRLMotionPlanning,Behzadan2021AdversarialRL}. This necessitates stabilized training paradigms, such as curriculum-based learning, to ensure robust policy convergence across heterogeneous pothole geometries.

This paper proposes a vehicle-dynamics-oriented hierarchical framework for coordinated pothole handling. By embedding impact-load limits and vertical-dynamics objectives into tactical decisions, the framework balances structural safety, ride comfort, and interactive traffic safety through coordinated lane-bounded lateral offsets and longitudinal speed regulation. The main contributions are:
\begin{enumerate}[label=(\arabic*)]
\item A vehicle-dynamics-oriented hierarchical framework that integrates physical impact-load limits and vertical-dynamics objectives into tactical decisions and generates kinematically feasible trajectories for downstream execution.
\item A Gymnasium-based multi-physics simulation testbed that unifies vehicle vertical dynamics, structural impact mechanics, and interactive car-following behavior. This platform enables unified, closed-loop assessment of conflicting driving objectives by translating physical impact boundaries into explicit state-reward formulations.
\item A curriculum-based maximum-entropy Soft Actor-Critic (SAC) training scheme to overcome sparse impact feedback and highly heterogeneous pothole geometries, ensuring robust policy convergence through progressive task-scaling.
\end{enumerate}

The remainder of this paper is organized as follows: Section~\ref{sec:literature_review} reviews related literature on vehicle control technologies for pothole-hazard driving and reinforcement learning applications. Section~\ref{sec:methodology} presents the proposed methodology, detailing the physics-informed impact mechanics model, the vehicle-dynamics-oriented hierarchical framework, and the curriculum learning strategy. Section~\ref{sec:experiment} describes the experimental setup. Section~\ref{sec:results} presents the experimental results, focusing on learning dynamics and emergent pothole-handling strategies. Section~\ref{sec:discussion} discusses policy generalization and study limitations. Finally, Section~\ref{sec:conclusion} summarizes the main findings and outlines directions for future research.

% ==========================================
% SECTION 2: LITERATURE REVIEW
% ==========================================
\section{Literature Review}
\label{sec:literature_review}

\subsection{Vehicle Control Technologies for Pothole-Hazard Driving}

Research into pothole hazards has evolved toward integrated vehicle control and planning methodologies that exploit preview data to reduce impact loads and ride discomfort while ensuring executable motion~\cite{chaaban2025end,JLRPotholeDetection,Yebes2021PotholeDetection}. At the system level, numerous studies have incorporated pothole detection and evasion into autonomous driving architectures, where perception units supply pothole localization along with geometric information, and subsequent modules devise feasible avoidance strategies~\cite{Yebes2021PotholeDetection,Abbasi2025LidarSegmentation,xiang2023multi}. These integrated systems bolster the viability of pothole-aware autonomy; however, many formulations predominantly address pothole handling as a geometric avoidance task. As a consequence, planning and control layers frequently overlook physically informed objectives and constraints that directly gauge transient wheel-pothole interactions and resultant vertical dynamics, which are crucial for evaluating structural risk and ride comfort during pothole encounters.

Within the control and motion-planning literature, pothole-impact mitigation has mainly evolved along two complementary directions: preview-based suspension control and longitudinal speed regulation. Preview-based suspension control exploits look-ahead road information to proactively adjust active or semi-active suspension responses, thereby reducing body vibration and wheel-load variation relative to purely reactive strategies~\cite{Theunissen2021PreviewReview}. In parallel, longitudinal speed regulation mitigates pothole-induced excitation by adapting the traversal speed before wheel entry, since the resulting vertical disturbance depends jointly on road geometry and vehicle speed~\cite{Sun2018OptimalDriving}. Building on these insights, recent studies have increasingly coupled preview suspension control with velocity planning to jointly balance ride comfort, longitudinal smoothness, travel efficiency, and impact mitigation~\cite{Wu2020SpeedSuspensionTVT,Liu2024PreviewSuspensionVelocity,Huang2023BioinspiredPreview}. However, these approaches remain centered on longitudinal mitigation of the previewed disturbance, with limited consideration of lane-bounded lateral maneuvering opportunities and interactive traffic constraints.

Beyond comfort-oriented control, pothole-aware planning now integrates safety considerations stemming from vehicle interactions. Recent research infers road-induced vibration from a leading vehicle's vertical response, enabling anticipatory speed planning while ensuring interaction safety~\cite{Zhang2023BumpyITSC}. Such advancements underscore that pothole-hazard driving transcends individual vehicle vibration management, evolving into an interactive planning and control challenge under conditions of limited observability.

Despite these advances, important limitations remain. First, most existing approaches treat vertical performance primarily as a low-level control objective and do not systematically elevate impact-load constraints and vertical-dynamics metrics to the tactical decision-making layer.  Second, because potholes are geometrically localized and wheel-trajectory dependent, lane-bounded lateral redundancy offers meaningful mitigation opportunities through bypassing or straddling maneuvers. Nevertheless, the majority of current methods focus predominantly on longitudinal optimization, leaving lateral repositioning under geometric and interaction-safety constraints comparatively underexplored. These gaps motivate a vehicle-dynamics-oriented decision-planning-control framework that jointly coordinates lateral offset selection and longitudinal speed regulation under physically interpretable impact-load and ride-comfort objectives.

\subsection{RL for Autonomous Driving and Vehicle Control}

Reinforcement learning has been extensively studied as a unifying framework for autonomous driving decision-making and vehicle control due to its capacity to optimize policies in nonlinear and stochastic environments through iterative interaction. By encoding multiple objectives within a structured reward design, RL enables coordinated optimization across safety, comfort, efficiency, and constraint satisfaction. Comprehensive surveys have categorized RL applications throughout the autonomy stack, including high-level behavior planning in interactive traffic, motion planning under complex kinematic and dynamic constraints, and low-level tracking and stabilization control~\cite{Wu2024RLBPSurvey,Aradi2022SurveyDRLMotionPlanning,hu2025survey,wang2024survey}. Collectively, these studies emphasize that RL is particularly advantageous in long-tailed or highly interactive scenarios where handcrafted rules are insufficient.

A parallel body of research has investigated RL for vehicle control problems characterized by strong dynamic coupling and parametric uncertainty. In such settings, classical controllers may exhibit sensitivity to modeling inaccuracies or varying operating conditions. Representative applications include suspension vibration mitigation, preview-based chassis control, and speed planning over rough or degraded road surfaces, all of which involve non-linear dynamics and competing performance objectives~\cite{Zhao2024ReachabilityGuidedRL,Wang2023RLMPC,Du2022RoughSpeedDRL}. Recent developments increasingly incorporate physics-guided and constraint-aware learning formulations, in which action spaces are restricted to physically meaningful control variables and safety-critical bounds are explicitly enforced to improve deployability and mitigate unsafe exploration~\cite{Zhao2024ReachabilityGuidedRL}.

Safety considerations have further driven advances in safe RL for autonomous driving, given that unconstrained exploration is incompatible with safety-critical domains. Recent surveys highlight constrained optimization, risk-sensitive objectives, and formal safety filters designed to enhance constraint satisfaction during both training and deployment~\cite{Inamdar2024SafeRLReview}. Hybrid architectures that integrate RL with model predictive control (MPC) have also emerged, leveraging MPC's structural guarantees and constraint-handling capabilities while allowing RL to adapt controller parameters across varying conditions~\cite{Wang2023RLMPC}. These approaches suggest that practical RL-based vehicle control benefits from structured formulations that explicitly unify physical constraints, interaction safety, and real-time feasibility within a closed evaluation loop framework.

In the context of pothole-hazard driving, the principal challenge lies not in the selection of RL per se, but in formulating an RL-compatible problem definition and training environment that accurately captures wheel-pothole impact mechanics, vertical vehicle dynamics, and interaction safety constraints within surrounding traffic. This perspective motivates the development of multi-physics simulation environments and maximum-entropy RL formulations capable of learning coordinated lateral-longitudinal strategies under sparse reward feedback and heterogeneous pothole geometries, while maintaining physically interpretable safety compliance~\cite{Testouri2025RoboCar,wang2024survey}. 

\subsection{Pothole Hazards as Frequent Pre-Crash Events}

Pothole-related roadway hazards constitute the second most frequently observed pre-crash event in the United States, which underscores their substantial contribution to crash risk~\cite{nhtsa_100carstudy,nhtsa_inattention,nhtsa_RealWorldBehaviors}. Large-scale naturalistic driving studies conducted by the National Highway Traffic Safety Administration (NHTSA) indicate roadway surface anomalies such as potholes are strongly associated with sudden steering inputs, abrupt uncontrolled braking and unanticipated path-crossing maneuvers that all increase the likelihood of driver loss-of-control events. These evasive actions, often executed under time-constrained conditions, destabilize the vehicle and increase the probability of both single-vehicle roadway departure crashes and multi-vehicle conflicts.

The high frequency of pothole-related pre-crash scenarios suggests that infrastructure condition is not only a maintenance concern but also a salient traffic safety issue. RL-based driver assistance systems offer a promising approach for mitigating pothole-related crashes by enabling adaptive, real-time responses to hazardous conditions. Unlike rule-based control strategies, RL algorithms can learn optimal braking and stabilizing policies through iterative interaction with diverse driving scenarios. RL agents can be trained to detect potholes and determine context-aware control actions, such as graded deceleration or lane-stabilizing steering adjustments that can minimize vehicle instability while preserving occupant safety and traffic flow. Reward functions can be explicitly designed to penalize excessive jerk, loss of traction, and other unsafe driver behaviors thereby encouraging smoother and more controlled braking maneuvers. When deployed as an advanced driver assistance technology, RL-based pothole mitigation systems have the potential not only to reduce the high incidence of loss-of-control crashes, but also reduce crash severity by proactively moderating vehicle speed before impact or poorly-controlled evasive maneuvers occur~\cite{nhtsa_RealWorldBehaviors}.

% ==========================================
% SECTION 3: METHODOLOGY
% ==========================================
\section{Methodology}
\label{sec:methodology}

\begin{figure*}[!t]
    \centering
    \includegraphics[width=0.9\textwidth]{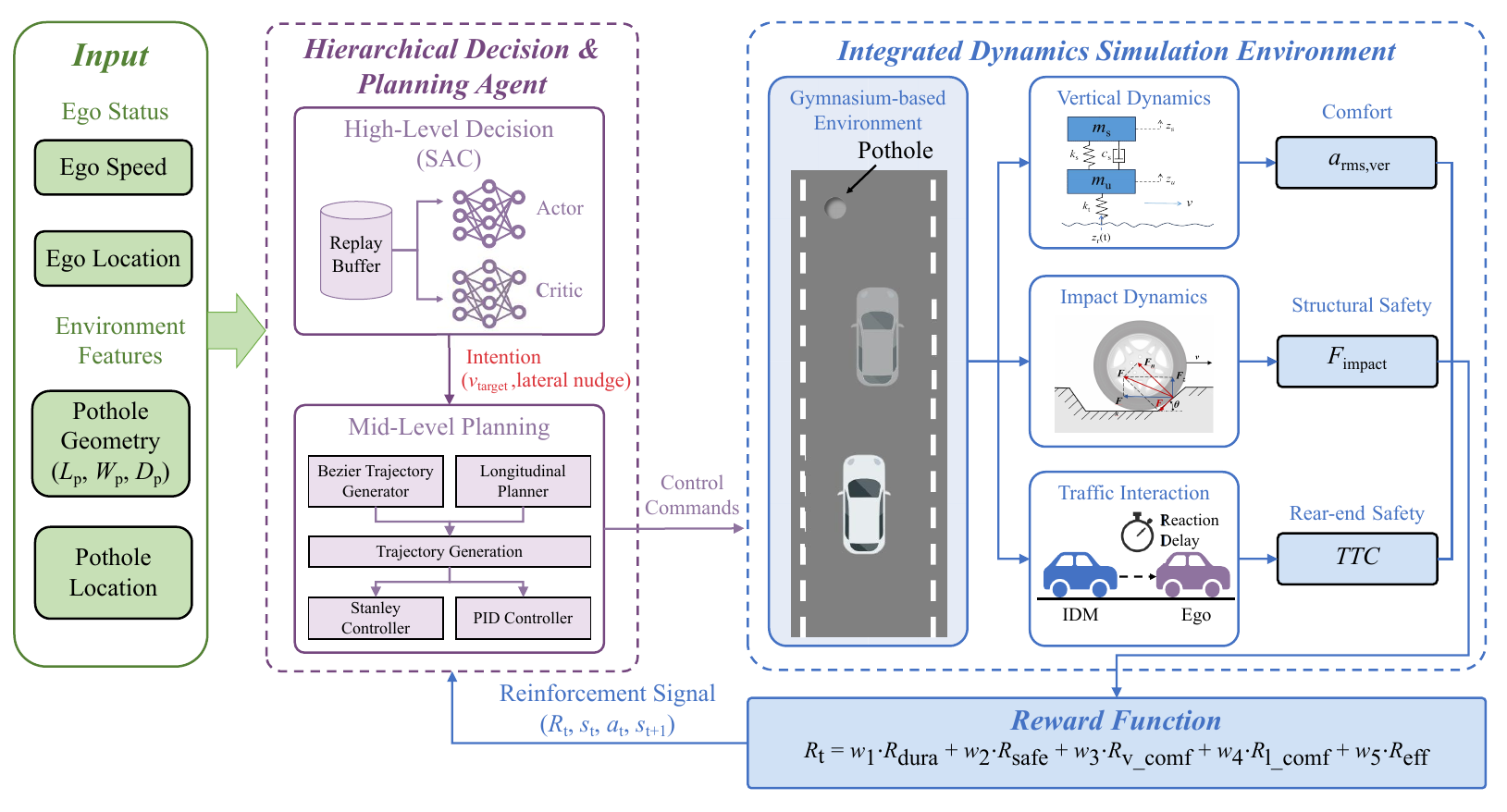}
    \caption{Overall framework of the proposed vehicle-dynamics-oriented hierarchical autonomy for pothole hazards.}
    \label{framework}
\end{figure*}

This section presents the proposed hierarchical framework for pothole-aware motion planning. As illustrated in Fig.~\ref{framework}, the framework integrates tactical decision-making with detailed modeling of vehicle-road interaction mechanics within a closed-loop, multi-physics simulation environment. The architecture simulates interactive traffic dynamics concurrently with mechanism-informed structural impacts to assess risks and define failure boundaries. Within this hierarchy, a high-level RL agent processes the scenario state and outputs a tactical action consisting of a target speed and a target lateral offset.

\subsection{Impact Mechanics Modeling}
\label{subsec:impact_mechanics}

To quantitatively evaluate the risk of structural damage during pothole traversal, we model the problem from a multi-body contact dynamics perspective rather than relying solely on classical vertical vibration theory. This formulation is necessary to capture the impulsive loads associated with tire and rim failures.

\subsubsection{Suspension Dynamics via Quarter-Car Model}
The vehicle's vertical dynamics are modeled using a 2-DOF quarter-car system, consisting of a sprung mass $m_s$ and an unsprung mass $m_u$. Assuming $z_s$ and $z_u$ denote the vertical displacements of the respective masses from their static equilibrium positions, the system's dynamic equations of motion are governed by:
\begin{equation}
\label{eq:quarter_car}
\begin{cases}
m_s \ddot{z}_s = -F_s, \\
m_u \ddot{z}_u = F_s - F_t + F_{imp,z},
\end{cases}
\end{equation}
where $F_s$ is the suspension force, $F_t$ is the tire-road interaction force, and $F_{imp,z}$ represents the explicit external impulsive force derived from the transient tire-pothole edge collision.

To accurately capture the nonlinear characteristics of the vehicle's vertical response under extreme excitations, the suspension force $F_s$ is formulated to include spring elasticity, asymmetric damping, and bump-stop nonlinearities:
\begin{equation}
\label{eq:suspension_force}
F_s = k_s (z_s - z_u) + F_{damp} + F_{bs}.
\end{equation}
Here, $k_s$ is the linear suspension stiffness. Real-world shock absorbers exhibit asymmetric characteristics; thus, the damping force $F_{damp}$ is modeled as a piecewise linear function depending on the suspension velocity $\Delta \dot{z} = \dot{z}_s - \dot{z}_u$:
\begin{equation}
F_{damp} = 
\begin{cases} 
c_{s,\mathrm{reb}} (\dot{z}_s - \dot{z}_u), & \text{if } \dot{z}_s - \dot{z}_u \ge 0, \\
c_{s,\mathrm{comp}} (\dot{z}_s - \dot{z}_u), & \text{if } \dot{z}_s - \dot{z}_u < 0,
\end{cases}
\end{equation}
where $c_{s,\mathrm{reb}}$ and $c_{s,\mathrm{comp}}$ are the rebound and compression damping coefficients, respectively. 

Furthermore, to prevent non-physical suspension penetration during severe pothole impacts, the bump-stop force $F_{bs}$ is activated when the suspension deflection $\Delta z = z_s - z_u$ exceeds the maximum physical travel limit $z_{\max}$:
\begin{equation}
F_{bs} =
\begin{cases} 
k_b ( (z_s - z_u) - z_{\max} ), & \text{if } z_s - z_u > z_{\max}, \\
k_b ( (z_s - z_u) + z_{\max} ), & \text{if } z_s - z_u < -z_{\max}, \\
0, & \text{otherwise},
\end{cases}
\end{equation}
where $k_b$ represents the high stiffness of the boundary limiters.

Finally, assuming the tire maintains continuous contact with the surface or experiences linear deformation, the tire force is modeled as:
\begin{equation}
F_t = k_t (z_u - z_r) + c_t (\dot{z}_u - \dot{z}_r),
\end{equation}
where $k_t$ and $c_t$ are the vertical stiffness and damping coefficient of the tire, and $z_r$ is the nominal road profile excitation. The parameters calibrated for the implemented quarter-car model are systematically detailed in Table~\ref{tab:veh_params}.

\subsubsection{Wheel--Pothole Impact Dynamics}

\begin{figure}[!t]
    \centering
    \includegraphics[width=0.9\columnwidth]{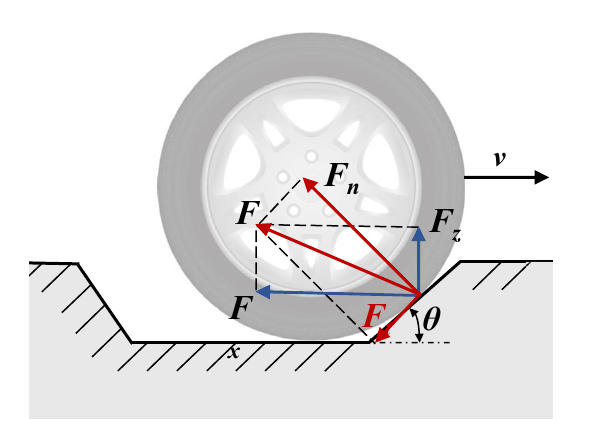}
    \caption{Proposed mechanism-informed model for tire--pothole edge impact analysis.}
    \label{fig:impact_mechanics}
\end{figure}

When the wheel encounters the leading edge of a pothole at high speed, a short-duration but high-magnitude impact load is generated. To capture this process, we adopt a reduced-order contact-impact formulation that accounts for the wheel's forward motion and the local wheel--edge contact geometry, as illustrated in Fig.~\ref{fig:impact_mechanics}.

At the onset of contact, the critical contact angle $\theta_c$ is determined by the pothole depth $D_p$ and the effective tire radius $R_t$:
\begin{equation}
\label{eq:impact_angle}
\theta_c = \arccos\left(1 - \frac{D_p}{R_t}\right),
\end{equation}
where $D_p$ denotes the pothole depth and $R_t$ denotes the effective tire radius.

The normal contact force is modeled using a penetration-based nonlinear contact law with a Hunt--Crossley-type velocity correction:
\begin{equation}
\label{eq:normal_force}
F_n = \max \left( K_c \delta^n \left(1 + \alpha_{hc} v_c\right),\, 0 \right),
\end{equation}
where $K_c$ is the effective contact stiffness, $\delta$ is the normal penetration depth, $n$ is the contact force exponent, $\alpha_{hc}$ is the velocity correction coefficient, and $v_c$ is the compressive normal relative velocity at the contact point.

Tangential interaction is modeled using Coulomb sliding friction:
\begin{equation}
\label{eq:friction_force}
F_f = \mu F_n,
\end{equation}
where $\mu$ is the sliding friction coefficient.

The resultant contact force is decomposed into longitudinal and vertical components according to the instantaneous contact angle $\theta$:
\begin{equation}
\label{eq:force_decomposition}
\begin{cases}
F_x = F_n \sin\theta + F_f \cos\theta, \\
F_z = F_n \cos\theta - F_f \sin\theta.
\end{cases}
\end{equation}

To quantify the severity of the edge strike, we define the peak resultant wheel-hub load during the impact interval as
\begin{equation}
\label{eq:peak_force}
F_{\text{peak}} = \max_{t \in [0,\,T_{\text{imp}}]} \sqrt{F_x^2(t) + F_z^2(t)},
\end{equation}
where $T_{\text{imp}}$ denotes the effective contact duration. This physics-based metric is used as the structural-safety indicator in the subsequent tactical decision-making process. The key parameters of the impact model are summarized in Table~\ref{tab:impact_params}.

\subsection{Hierarchical Planning Framework}
\label{subsec:framework}

\begin{figure}[!t]
    \centering
    \includegraphics[width=\columnwidth]{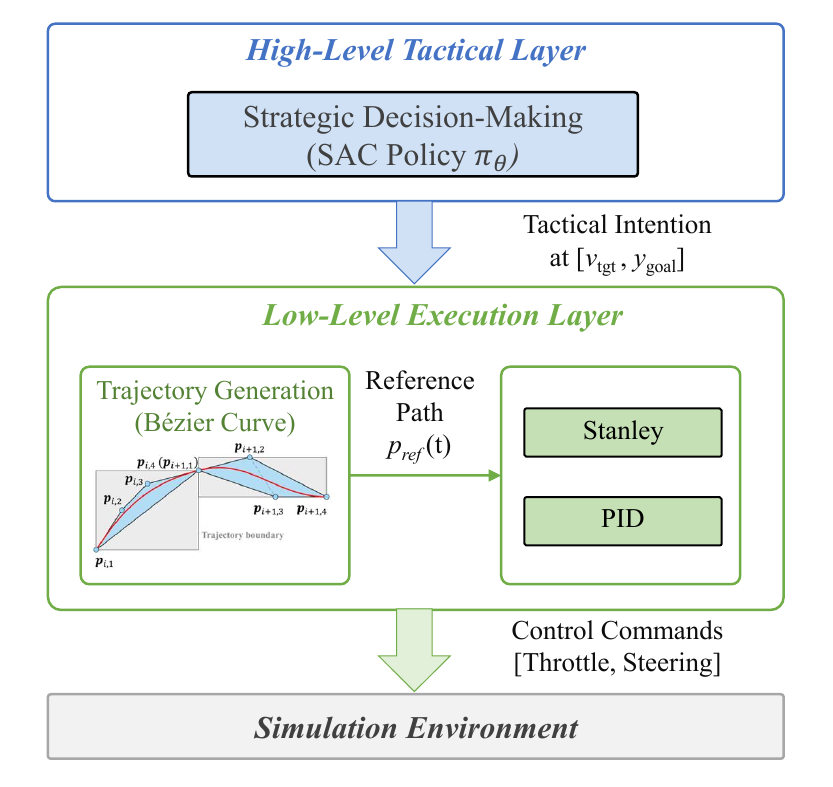}
    \caption{Architecture of the hierarchical decision-making and low-level execution layers.}
    \label{fig:hierarchical_agent}
\end{figure}

The proposed framework adopts a \emph{vehicle-dynamics-oriented} hierarchical architecture that separates tactical action generation from trajectory realization. The internal architecture of the planning agent is illustrated in Fig.~\ref{fig:hierarchical_agent}.

As shown in Fig.~\ref{fig:hierarchical_agent}, the system operates through a closed-loop interaction between three core modules: a high-level strategic layer, a low-level execution layer, and a physics-informed simulation environment.

Unlike conventional end-to-end continuous control where the policy operates at a high frequency (e.g., 10~Hz), our tactical decision-making is formulated as an \emph{episodic, single-step optimization problem}. Upon previewing a hazard, the agent selects a one-step tactical decision, and the episode terminates after the low-level controllers execute the maneuver and physical outcomes are evaluated. Consequently, the process is modeled as a continuous-action contextual bandit problem, or equivalently, a single-step Markov Decision Process (MDP) defined by the tuple $(\mathcal{S}, \mathcal{A}, \mathcal{R})$ without a temporal discount factor.
\begin{itemize}
    \item \textbf{State Space $\mathcal{S}$}: Encodes the initial mission context $s_0$ at the exact moment the pothole is previewed, including the ego vehicle's kinematics (e.g., initial velocity $v_{ego}$), the pothole's geometric attributes (length $L_p$, width $W_p$, depth $D_p$), and the relative state of surrounding traffic.
    \item \textbf{Action Space $\mathcal{A}$}: A continuous space representing the one-step tactical action $\mathbf{u} = [v_{\text{tgt}},\, y_{\text{tgt}}]$. Here, $v_{\text{tgt}}$ denotes the target traversal velocity, and $y_{\text{tgt}}$ specifies the target lateral offset within the lane boundaries.
    \item \textbf{Reward $\mathcal{R}$}: Maps the selected tactical action and its resulting episode outcome to a scalar return that encodes trade-offs among passage efficiency, rear-end safety, ride comfort, and structural durability. 
\end{itemize}

The high-level policy $\pi_\theta(\mathbf{u}\mid s_0)$ is optimized using the Soft Actor-Critic (SAC) algorithm. Although typically applied to sequential tasks, SAC is particularly advantageous in this single-step context due to its maximum-entropy framework. In our highly constrained environment, naive optimization frequently causes the policy to collapse into sub-optimal local minima (e.g., passive hard braking without lateral avoidance). SAC's entropy regularization promotes robust exploration in the continuous tactical action space. Under this single-step formulation, the critic is trained from the terminal reward of the selected tactical action, while the entropy term helps preserve exploration diversity. The chosen tactical actions are subsequently executed by the low-level layer, where a quintic B\'ezier generator constructs a smooth path and robust controllers (Stanley and PID) generate actuator commands to interact with the environment.

\subsection{Reward Function}
\label{subsec:reward}

The reward function is designed to balance passage efficiency, rear-end interaction safety, ride comfort, and structural durability. Because the framework operates as a single-step decision process and the low-level controller is designed to complete the maneuver, premature termination cannot be exploited to avoid penalties. Accordingly, the episode return is formulated as the negative weighted sum of physically grounded penalty terms.
The total reward $R_{\text{total}}$ is defined as
\begin{equation}
\label{eq:total_reward}
R_{\text{total}} = W_{\text{eff}}\, r_{\text{eff}} + W_{\text{rear}}\, r_{\text{rear}} + W_{\text{comf}}\, r_{\text{comf}} + W_{\text{dura}}\, r_{\text{dura}},
\end{equation}
where $W_{\text{eff}}, W_{\text{rear}}, W_{\text{comf}}, W_{\text{dura}} > 0$ are scalar weights that prioritize competing objectives, and each $r \le 0$ is a physically grounded penalty term computed over the episode rollout.

\subsubsection{Passage Efficiency}
To discourage overly conservative behaviors such as unnecessary stopping or crawling, the agent is penalized for the squared \emph{relative} delay.
To support magnitude normalization across heterogeneous scenarios, the efficiency penalty is normalized by the reference travel time:
\begin{equation}
\label{eq:reward_eff}
r_{\text{eff}} = - \left( \frac{\max(0,\, T_{\text{act}} - T_{\text{ref}})}{T_{\text{ref}}} \right)^2,
\end{equation}
where $T_{\text{act}}$ is the realized travel time of the episode and $T_{\text{ref}}$ denotes the reference travel time required to traverse the segment at the initial speed limit.

\subsubsection{Traffic Safety}
Rear-end collision risk induced by aggressive braking is quantified using the inverse time-to-collision ($\mathrm{iTTC}$) relative to the following vehicle.
The traffic-safety penalty is accumulated over the episode rollout as
\begin{equation}
\label{eq:reward_safety}
r_{\text{rear}} = - \sum_{k=1}^{N} \left( \max\!\left(0,\, \mathrm{iTTC}_k - \mathrm{iTTC}_{\text{thr}}\right) \right)^2.
\end{equation}
Following empirical car-following safety analyses~\cite{hussain2025empirical}, we set $\mathrm{iTTC}_{\text{thr}} = 0.33\,\mathrm{s}^{-1}$ (corresponding to $\mathrm{TTC}=3.0$~s) to maintain a conservative car-following safety margin.

\subsubsection{Ride Comfort}
Ride comfort is evaluated using ISO-weighted sprung-mass accelerations in the vertical directions.
To accurately reflect human perception, the raw accelerations are filtered using the ISO~2631-1 frequency weighting, yielding ISO-weighted sprung-mass accelerations $a_{z,w}$ and $a_{x,w}$.
The unified comfort penalty is defined as
\begin{equation}
\label{eq:reward_ride}
r_{\text{comf}} = - \left( \mathrm{RMS}(a_{z,w})^2 + w_{\text{lon}}\, \mathrm{RMS}(a_{x,w})^2 \right),
\end{equation}
where $w_{\text{lon}}$ balances longitudinal braking intensity against the primary vertical suspension response.
Here $\mathrm{RMS}(x)=\sqrt{\frac{1}{N}\sum_{k=1}^{N}x_k^2}$ is computed over episode samples, and the ISO~2631-1 weighting is realized in discrete time at the simulator sampling interval $\Delta t$.

\subsubsection{Structural Durability}
This component explicitly enforces the vehicle-dynamics-oriented safety constraint, targeting the risk of wheel and rim damage during edge strikes.
The penalty activates when the peak resultant force $F_{\text{peak}}$ exceeds a warning threshold $F_{\text{warn}}$:
\begin{equation}
\label{eq:reward_struct}
r_{\text{dura}} = - \left( \frac{\max(0,\, F_{\text{peak}} - F_{\text{warn}})}{F_{\text{lim}} - F_{\text{warn}}} \right)^2,
\end{equation}
where $F_{\text{lim}}$ denotes the catastrophic rim failure limit.
Based on full-vehicle impact experiments~\cite{Zhang2020TireObstacle}, we set $F_{\text{warn}} = 10.6\,\mathrm{kN}$ to represent the onset of potential mechanical damage and $F_{\text{lim}} = 15.15\,\mathrm{kN}$ as the hard physical boundary.

\subsubsection{Reward Calibration}
The objective weights are empirically tuned following the principle of magnitude normalization, such that the \emph{unweighted} penalty components ($r_{\text{eff}}, r_{\text{rear}}, r_{\text{comf}}, r_{\text{dura}}$) remain within the same order of magnitude under typical pothole-encounter scenarios.
This normalization prevents any single metric with inherently large numerical scale from dominating gradient updates.
Building upon this baseline, scalar weights are further adjusted to reflect safety-first priorities in mixed traffic, where durability and comfort are prioritized over efficiency.
The detailed parameter configuration is summarized in Table~\ref{tab:reward_weights}.

\begin{table}[!t]
\centering
\caption{Configuration of reward function weights.}
\label{tab:reward_weights}
\renewcommand{\arraystretch}{1.2}
\begin{tabular}{llc}
\toprule
\textbf{Objective Component} & \textbf{Symbol} & \textbf{Value} \\
\midrule
Passage Efficiency     & $W_{\text{eff}}$   & 1.0 \\
Traffic Safety         & $W_{\text{rear}}$  & 10.0 \\
Unified Ride Comfort   & $W_{\text{comf}}$  & 10.0 \\
\quad \textit{Longitudinal sub-weight} & $w_{\text{lon}}$ & 1.0 \\
Structural Durability  & $W_{\text{dura}}$  & 20.0 \\
\bottomrule
\end{tabular}
\end{table}

\subsection{Curriculum Learning Strategy}
\label{subsec:curriculum}

In the high-speed pothole-handling task, the agent faces a highly non-convex reward landscape caused by the coupling among impact safety, ride comfort, traffic interaction, and lane-bounded maneuvering. Standard reinforcement learning often struggles in such settings because useful policies are difficult to discover through unconstrained exploration alone, especially when severe impact events induce sparse and high-variance penalties~\cite{narvekar2020curriculum,lin2025survey}.

To improve training stability, we employ a two-stage curriculum learning strategy that gradually increases task complexity, following the general principle of progressive task sequencing in curriculum-based reinforcement learning~\cite{narvekar2020curriculum,lin2025survey}. The core idea is to first establish a reliable longitudinal protection policy, and then introduce more diverse pothole geometries that require coordinated lateral--longitudinal decision-making.

\textbf{Stage I: Longitudinal anchoring.}
In the first stage, the agent is primarily exposed to large and difficult-to-bypass potholes. This stage simplifies the decision structure and encourages the policy to learn the dominant relationship between speed reduction and impact-load mitigation, thereby establishing a stable safety-oriented baseline.

\textbf{Stage II: Lateral--longitudinal coordination.}
After the initial anchoring stage, the training distribution is expanded to include a mixture of large unavoidable potholes, wheel-path potholes, and randomly located moderate potholes. This stage requires the agent to jointly optimize target speed and lateral offset, enabling the emergence of coordinated avoidance, straddling, and compliant traversal behaviors.

In the implemented training schedule, Stage I occupies the first 300 episodes. From episode 301 onward, the training distribution switches to a mixed regime consisting of 30\% large unavoidable potholes, 40\% wheel-path potholes, and 30\% randomly located potholes. The complete training and interaction procedure, including the stage-dependent scenario sampling strategy, is summarized in Algorithm~\ref{alg:training_flow}. This schedule-based curriculum avoids abrupt exploration over the full task space at the start of training and improves convergence robustness in the coupled multi-objective environment.

\begin{algorithm}[!t]
\caption{Training and interaction procedure of the proposed hierarchical framework}
\label{alg:training_flow}
\begin{algorithmic}[1]
\Require Replay buffer $\mathcal{B}$; total episodes $M$.
\Ensure Trained tactical policy $\pi_\theta$.
\State Initialize stochastic policy $\pi_\theta$ and value networks $Q_{\phi_1}, Q_{\phi_2}$.
\For{episode $e = 1$ to $M$}
    \If{$e \leq 300$}
        \State Sample a training scenario from the Stage-I distribution.
    \Else
        \State Sample a training scenario from the Stage-II mixed distribution.
    \EndIf
    
    \State Observe initial preview state $s_0$ (vehicle kinematics, pothole geometry, and traffic context).
    
    \State \textbf{Tactical decision (single-step high-level):}
    \State Generate tactical action $\mathbf{u}_e = [v_{tgt},\, y_{tgt}] \sim \pi_\theta(\cdot \mid s_0)$.
    
    \State \textbf{Execution and physical interaction (continuous low-level):}
    \State Generate a reference path using a quintic B\'ezier curve based on $\mathbf{u}_e$.
    \State Execute closed-loop tracking via Stanley and PID controllers until the hazard is passed.
    \State Compute the episodic physical outcomes, including peak impact force $F_{\text{peak}}$ and traffic-safety indicators.
    
    \State \textbf{Policy evaluation and update:}
    \State Calculate the episodic return $R_{\text{total}}$ using (\ref{eq:total_reward}).
    \State Store terminal transition $(s_0, \mathbf{u}_e, R_{\text{total}}, s_{\text{term}}, d{=}1)$ in replay buffer $\mathcal{B}$.
    \State Update $\pi_\theta$ and $Q_{\phi_{1,2}}$ using the SAC objective for terminal one-step returns.
\EndFor
\end{algorithmic}
\end{algorithm}

% ---------- Section 4 Experimental Setup ----------
\section{Experimental Setup}
\label{sec:experiment}

This section describes the experimental setup used to train and evaluate the proposed vehicle-dynamics-oriented tactical decision-making method for pothole handling. We first introduce a Gymnasium-based multi-physics environment that integrates quarter-car vertical dynamics, tire--pothole impact interaction, and reaction-time delayed car-following behavior. We then present the empirical parameterization adopted for pothole geometry sampling. Baseline strategies are summarized in Section~\ref{subsec:baselines}, and SAC implementation details are provided in Section~\ref{subsec:rl_impl}.

\subsection{Integrated Multi-Physics Simulation Environment}
\label{subsec:sim_platform}

The environment follows the Gymnasium interface to provide a standardized agent--environment interaction loop for reinforcement learning~\cite{towers2024gymnasium}. Each episode is initialized such that the ego vehicle approaches a pothole with an initial speed uniformly sampled from $[60,120]$~km/h, and a stochastic preview distance sampled from $[10,100]$~m to emulate variability in detectability.

\paragraph{State and action interface}
At each decision step, the agent observes a state vector that encodes (i) the current longitudinal speed, (ii) the preview distance to the pothole, and (iii) the pothole geometry and lateral alignment relative to the ego lane. The agent outputs a two-dimensional continuous tactical action $\mathbf{u}=[v_{\mathrm{tgt}},\, y_{\mathrm{tgt}}]$, where $v_{\mathrm{tgt}}$ denotes the target speed and $y_{\mathrm{tgt}}$ denotes the target lateral offset within the lane bounds. Here, $y_{\max}$ represents the maximum admissible lateral offset determined by the lane geometry and safety margin. This design keeps the high-level policy focused on tactical decisions, while low-level deterministic tracking ensures smooth execution.

\paragraph{Vertical dynamics}
The ego vehicle's vertical dynamics are modeled by a 2-DOF quarter-car system, capturing sprung mass and unsprung mass dynamics for comfort evaluation and impact transient coupling.
Key parameters used in the simulator are summarized in Table~\ref{tab:veh_params}.

\begin{table}[!t]
\centering
\caption{Quarter-car model parameters.}
\label{tab:veh_params}
\renewcommand{\arraystretch}{1.2}
\begin{tabular}{lcc}
\toprule
\textbf{Parameter} & \textbf{Symbol} & \textbf{Value} \\
\midrule
Sprung mass & $m_s$ & $350$ kg \\
Unsprung mass & $m_u$ & $45$ kg \\
Suspension stiffness & $k_s$ & $3.0\times10^4$ N/m \\
Tire stiffness & $k_t$ & $2.0\times10^5$ N/m \\
Tire damping & $c_t$ & $200$ N$\cdot$s/m \\
Compression damping & $c_{s,\mathrm{comp}}$ & $1000$ N$\cdot$s/m \\
Rebound damping & $c_{s,\mathrm{reb}}$ & $3000$ N$\cdot$s/m \\
Maximum suspension travel & $z_{\max}$ & $0.10$ m \\
Bump-stop stiffness & $k_b$ & $5.0\times10^5$ N/m \\
\bottomrule
\end{tabular}
\end{table}

\paragraph{Mechanism-informed tire--pothole impact coupling}
To explicitly capture edge-strike transients beyond acceleration-only surrogates, the simulator invokes the mechanism-informed tire--pothole impact model introduced in Section~\ref{subsec:impact_mechanics}. 
For each episode, the impact solver returns the peak resultant impact load, which is used by the structural-safety objective and enables constraint-aware tactical learning under pothole hazards. The main parameters used in the impact model are summarized in Table~\ref{tab:impact_params}.

\begin{table}[!t]
\centering
\caption{Parameters used in the wheel--pothole impact model.}
\label{tab:impact_params}
\renewcommand{\arraystretch}{1.2}
\begin{tabular}{lcc}
\toprule
\textbf{Parameter} & \textbf{Symbol} & \textbf{Value} \\
\midrule
Effective tire radius & $R_t$ & $0.31~\mathrm{m}$ \\
Normal contact stiffness & $K_c$ & $2.0\times10^6~\mathrm{N\,m}^{-n}$ \\
Contact force exponent & $n$ & $1.0$ \\
Velocity correction coefficient & $\alpha_{hc}$ & $0.25$ \\
Sliding friction coefficient & $\mu$ & $0.80$ \\
Equivalent wheel mass & $m_w$ & $50~\mathrm{kg}$ \\
\bottomrule
\end{tabular}
\end{table}

\paragraph{Traffic interaction with reaction-time constraints}
Interactive traffic is modeled by a following vehicle governed by the Intelligent Driver Model (IDM)~\cite{treiber2000congested}. To reflect realistic rear-end pressure during pothole handling, the follower is initialized with short time headways consistent with naturalistic observations~\cite{Zhu2016}.
Moreover, a perception--reaction delay is incorporated to avoid idealized instantaneous responses; we follow empirical reaction-time ranges in car-following studies~\cite{ChangChon2005ReactionTime}.
Table~\ref{tab:idm_params} reports the IDM and reaction-time settings used for simulation.

\begin{table}[!t]
\centering
\caption{Follower IDM and reaction-time parameters.}
\label{tab:idm_params}
\small
\setlength{\tabcolsep}{5pt}
\begin{tabular}{lcc}
\toprule
\textbf{Parameter} & \textbf{Symbol} & \textbf{Setting} \\
\midrule
Initial time headway & $T_{\mathrm{init}}$ & $\mathcal{U}(0.5,\,1.2)~\mathrm{s}$ \\
Desired time headway (IDM) & $T$ & $\mathcal{U}(0.5,\,T_{\mathrm{init}})~\mathrm{s}$ \\
Desired speed & $v_0$ & $\mathcal{U}(33,\,36)~\mathrm{m/s}$ \\
Max acceleration & $a$ & $1.4~\mathrm{m/s^2}$ \\
Comfortable deceleration & $b$ & $2.0~\mathrm{m/s^2}$ \\
Hard braking limit & $b_{\max}$ & $\mathcal{U}(4,\,6)~\mathrm{m/s^2}$ \\
Reaction time & $\tau_r$ & $\mathcal{U}(0.8,\,1.2)~\mathrm{s}$ \\
Integration step (traffic) & $\Delta t$ & $0.02~\mathrm{s}$ \\
\bottomrule
\end{tabular}
\end{table}

\subsection{Empirical Parameterization of Pothole Geometries}
\label{subsec:pothole_modeling}

\begin{figure}[!t]
    \centering
    \subfloat[RSRD samples.]{\includegraphics[width=0.48\columnwidth]{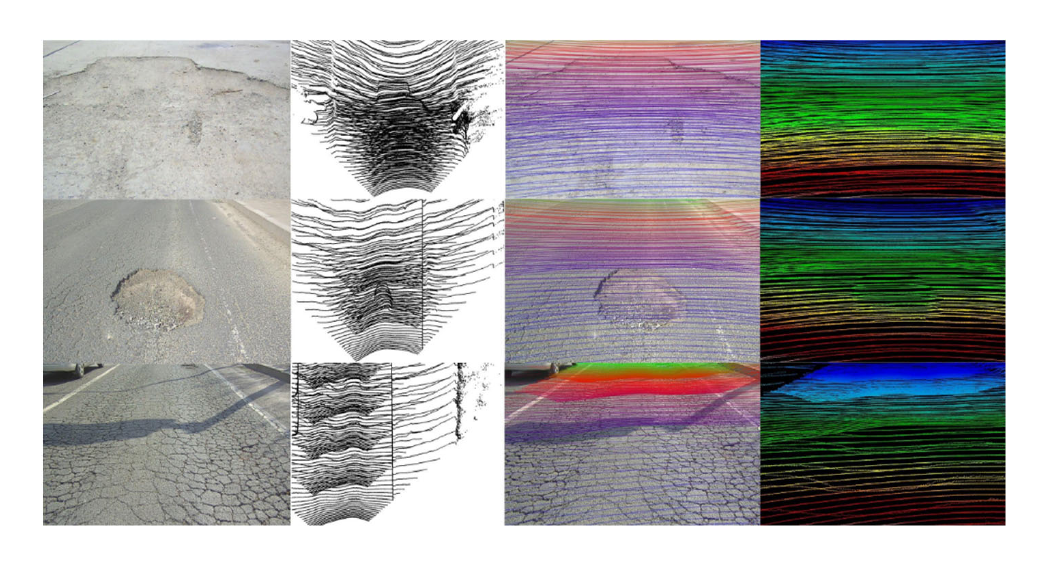}\label{fig:rsrd_sample}}
    \hfill
    \subfloat[In-house dataset samples.]{\includegraphics[width=0.48\columnwidth]{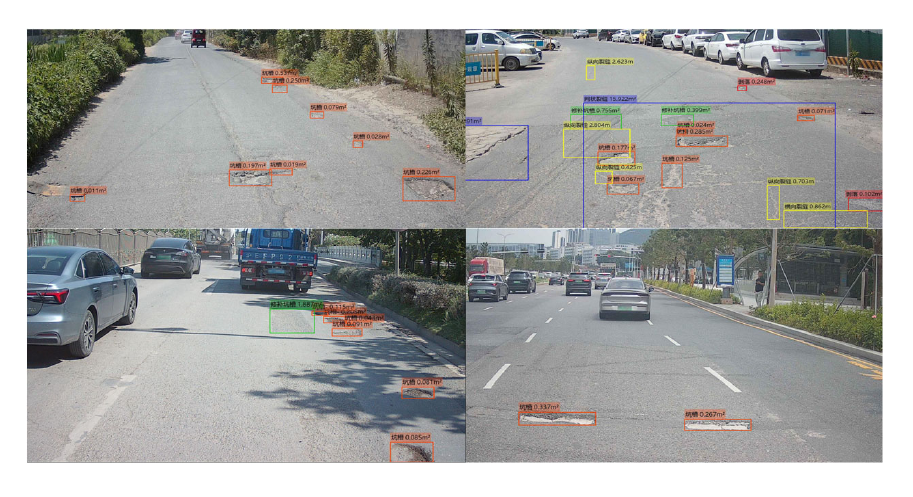}\label{fig:private_sample}}
    \caption{Representative pothole samples from the two data sources used for geometry parameterization.}
    \label{fig:pothole_datasets}
\end{figure}

Scenario generation is based on empirical pothole measurements to ensure realistic defect severities in both training and evaluation.
Each pothole is parameterized by its geometric dimensions: length $L_p$, width $W_p$, and depth $D_p$.
These parameters are estimated from two independent data sources (Fig.~\ref{fig:pothole_datasets}), including the Road Surface Reconstruction Dataset (RSRD)~\cite{Zhao2024RSRDScientificData} and an in-house dataset collected from multiple road segments across different regions.
In this study, the two sources play an equal role: both are used to statistically characterize pothole size distributions and enlarge the coverage of observed defect geometries, rather than introducing appearance- or texture-level variation.

Aggregating the measurements yields a long-tailed severity profile: most potholes are of moderate size, while a non-negligible subset exhibits large width and depth that may induce critical edge-strike loads.
Accordingly, we sample $L_p \in [0.2, 1.0]$~m, $W_p \in [0.3, 1.0]$~m, and $D_p \in [0.03, 0.15]$~m to construct realistic yet diverse pothole hazards, together with the preview-distance variability described in Section~\ref{subsec:sim_platform}.
This data-driven parameterization enables controlled scenario diversity for policy learning and evaluation.

\subsection{Baseline Strategies}
\label{subsec:baselines}

To evaluate the effectiveness of the proposed method, we benchmark against four representative strategies spanning passive execution, longitudinal-only learning and optimization, and geometry-driven lateral avoidance.

\begin{enumerate}
    \item \textbf{Constant Speed (CS):} A passive baseline that maintains the initial speed and lane-center trajectory throughout the scenario, providing a reference for unmitigated pothole traversal.
    
    \item \textbf{RL-based Speed Control (RL-SC):} Adapted from~\cite{Du2022RoughSpeedDRL}, this baseline learns longitudinal speed regulation to reduce vertical excitation and energy consumption, but does not use lateral avoidance.
    
    \item \textbf{Optimization-based Speed Planning (Opt-SP):} Based on~\cite{Zhang2023BumpyITSC}, this baseline optimizes a longitudinal speed profile under a lane-keeping assumption to mitigate road-induced excitation, but does not use lateral maneuvering space.
    
    \item \textbf{Heuristic Avoidance System (HAS):} Implemented following~\cite{chaaban2025end}, this baseline applies geometry-based clearance rules to generate lateral maneuvers. It bypasses the pothole when feasible and otherwise decelerates, without explicitly considering impact-load constraints.
\end{enumerate}

\subsection{Reinforcement Learning Implementation Details}
\label{subsec:rl_impl}

The proposed method is trained using Soft Actor-Critic (SAC), which serves as the optimizer for the single-step tactical policy.
Table~\ref{tab:sac_params} summarizes the main hyperparameters used in training. Unless otherwise stated, the results in Section~\ref{sec:results} are averaged over five random seeds.

\begin{table}[!t]
\centering
\caption{Main hyperparameters used for SAC training.}
\label{tab:sac_params}
\small
\setlength{\tabcolsep}{4pt}
\renewcommand{\arraystretch}{1.15}
\begin{tabular}{lcc}
\toprule
\textbf{Hyperparameter} & \textbf{Symbol} & \textbf{Value} \\
\midrule
Actor learning rate      & $\eta_{\pi}$      & $3\times10^{-4}$ \\
Critic learning rate     & $\eta_{Q}$        & $3\times10^{-4}$ \\
Target network update rate & $\tau$          & $0.005$ \\
Batch size               & $B$               & $64$ \\
Replay buffer size       & $|\mathcal{B}|$   & $1\times10^{5}$ \\
Updates per episode      & --                & $8$ \\
Warm-up episodes         & --                & $200$ \\
Entropy temperature      & $\alpha$          & auto ($\mathcal{H}_{\mathrm{tgt}}=-2$) \\
Hidden width             & --                & $256$ \\
\bottomrule
\end{tabular}
\end{table}

% ---------- Section 5 Results and Discussion ----------
\section{Experimental Results}
\label{sec:results}

This section reports the experimental results and analyzes the performance of the proposed method in pothole-handling scenarios. We first examine the learning dynamics to assess the proposed two-stage training strategy (Section~\ref{subsec:learning_dynamics}). We then compare the proposed method with representative baselines under factorized scenario settings (Section~\ref{subsec:comparative_analysis}), followed by representative case studies that illustrate the learned decision logic (Section~\ref{subsec:case_studies}). All learning-based methods use the training hyperparameters summarized in Table~\ref{tab:sac_params}, and the reported results are aggregated over multiple random seeds.

% ---------- Section 4.1 Learning dynamics and convergence ----------
\subsection{Learning Dynamics and Convergence}
\label{subsec:learning_dynamics}

\begin{figure}[!t]
    \centering
    \includegraphics[width=\columnwidth]{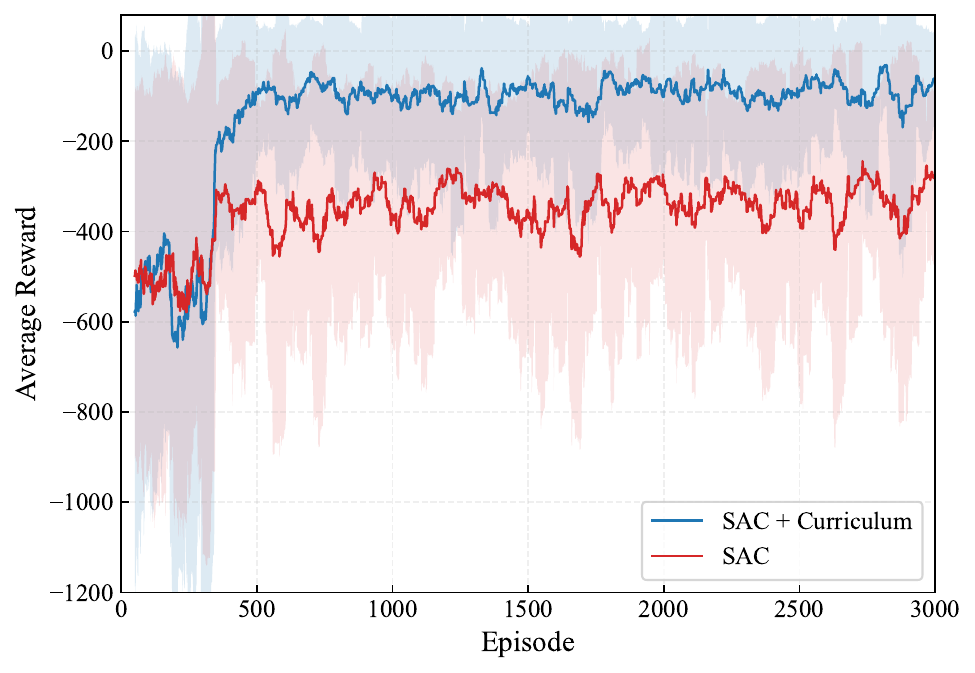}
    \caption{Training curves of episodic return. Solid lines denote the mean over five random seeds and shaded regions indicate the corresponding variability.}
    \label{fig:learning_curve}
\end{figure}

We examine the learning dynamics to assess the role of the proposed two-stage training strategy in stabilizing policy optimization under coupled safety, comfort, and interaction objectives. Specifically, we compare the proposed two-stage training scheme with a vanilla SAC variant trained under the same hyperparameter setting but without curriculum staging.
Fig.~\ref{fig:learning_curve} plots the episodic return during training, where solid lines denote the mean over five random seeds and the shaded regions indicate the corresponding variability.

The vanilla SAC baseline shows persistently higher return variance and a lower final return plateau.
This pattern is consistent with sparse, high-magnitude penalties induced by rare but severe events, such as excessive impact loads or rear-end conflicts, which amplify gradient noise and destabilize value estimation during training.

In contrast, the proposed staged training yields a faster rise in return and markedly reduced variability across seeds.
In Stage~1, the agent is exposed to simplified conditions that emphasize the dominant relationship between longitudinal speed regulation and impact-load mitigation, allowing it to learn a reliable baseline for satisfying the structural-safety constraint.
In Stage~2, more challenging lateral maneuvering and interactive constraints are introduced, which initially causes a transient fluctuation in return as the policy adapts to the expanded feasible set.
After adaptation, the learning curve stabilizes with substantially lower variance, indicating that the staged strategy improves optimization stability and leads to more consistent convergence in the coupled multi-objective environment.

% ---------- Section 4.2 Comparative performance analysis ----------
\subsection{Comparative Performance Analysis}
\label{subsec:comparative_analysis}

To evaluate the robustness of the proposed framework, we use a stratified sampling protocol to generate evaluation episodes. This section outlines the experimental design, defines the performance metrics, and presents a quantitative comparison against the baseline strategies.

\subsubsection{Experimental Design and Metrics}
\label{subsubsec:exp_design}

\textbf{Experimental Protocol:} Evaluation scenarios are constructed using a stratified factorial protocol over three key dimensions:
(i) \emph{Preview distance} (Short: 10--30~m; Medium: 30--60~m; Long: 60--100~m);
(ii) \emph{Pothole depth} (Shallow: 3--8~cm; Deep: 8--15~cm); and
(iii) \emph{Avoidability} (Avoidable vs.\ Unavoidable, determined by whether a feasible geometric solution exists within the lane boundaries). This yields $3 \times 2 \times 2 = 12$ core scenario groups. For each group, we simulate 20 independent trials with randomized traffic behaviors (IDM parameters) and ego-vehicle initial speeds ($v_{init} \in [60, 120]$~km/h), resulting in a total of 240 evaluation episodes per method.

\textbf{Performance Metrics:} To provide a multidimensional assessment of the tactical decision-making capability, we report five episode-level metrics:
\begin{itemize}
    \item \textbf{Avoidance rate (\%):} The percentage of episodes in which the tire footprint does not intersect the pothole polygon, resulting in zero impact force.
    \item \textbf{Structural safety rate (\%):} The percentage of episodes where the peak resultant impact force remains within the rim safety limit ($F_{\text{peak}} \le 15.15\,\mathrm{kN}$). This metric accounts for both successful avoidance and compliant traversal.
    \item \textbf{Collision rate (\%):} The percentage of episodes involving a rear-end collision with the following vehicle (gap $\le 0$).
    \item \textbf{Travel time (s):} The duration required for the ego vehicle to traverse the 200-meter road segment.
    \item \textbf{Vertical / Longitudinal Acceleration RMS ($\mathrm{m/s^2}$):} The Root Mean Square (RMS) of the sprung mass vertical acceleration ($a_z$) and longitudinal acceleration ($a_x$), characterizing ride comfort and braking intensity, respectively.
\end{itemize}

\subsubsection{Emergent Pothole-Handling Strategies}

\begin{figure*}[!t]
    \centering
    \subfloat[]{\includegraphics[width=0.9\textwidth]{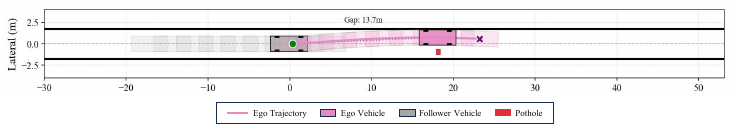}\label{fig:avoid_left}}
    
    \vspace{1em}
    
    \subfloat[]{\includegraphics[width=0.9\textwidth]{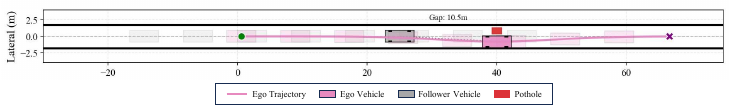}\label{fig:avoid_right}}
    
    \vspace{1em}
    
    \subfloat[]{\includegraphics[width=0.9\textwidth]{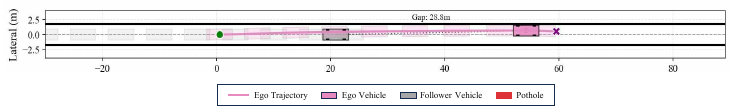}\label{fig:avoid_middle}}
    
    \caption{Visualization of emergent in-lane pothole avoidance behaviors learned by the proposed method. (a) Left nudge. (b) Right nudge. (c) Straddling (center pass).}
    \label{fig:avoidance_methods}
\end{figure*}

Before analyzing aggregated statistics, we examine the physical validity of the learned policy. As illustrated in Fig.~\ref{fig:avoidance_methods}, the proposed method demonstrates three distinct in-lane handling strategies: \emph{left nudge}, \emph{right nudge}, and \emph{straddling} (center pass). These behaviors are not manually specified, but instead emerge from the reward design, and provide the kinematic basis for the quantitative performance improvements reported below.

\subsubsection{Quantitative Results}

We compare the proposed method with Constant Speed (CS), RL-based Speed Control (RL-SC), Optimization-based Speed Planning (Opt-SP), and Heuristic Avoidance System (HAS). The statistical results are categorized into overall performance (Table~\ref{tab:all}), avoidable scenarios (Table~\ref{tab:avoidable}), and unavoidable scenarios (Table~\ref{tab:unavoidable}).

% 定义辅助命令

% =========================
% Table 1: All Scenarios
% =========================
\begin{table*}[!t]
\centering
\caption{Performance Statistics Across All 240 Evaluation Episodes}
\label{tab:all}
\renewcommand{\arraystretch}{1.2} % 稍微增加行高，让表格更美观
\footnotesize 
\setlength{\tabcolsep}{6pt} % 跨双栏后可以适当增大列间距
\begin{tabularx}{0.95\textwidth}{l | YYYYYY} % 添加了竖线 |，宽度设为 0.95 倍页面宽度
\toprule
\textbf{Method} & 
\makecell[c]{\textbf{Avoidance}\\\textbf{Rate (\%)}} & 
\makecell[c]{\textbf{Travel}\\\textbf{Time (s)}} & 
\makecell[c]{\textbf{Structural Safety}\\\textbf{Rate (\%)}} & 
\makecell[c]{\textbf{Collision}\\\textbf{Rate (\%)}} & 
\makecell[c]{\textbf{Vertical}\\\textbf{RMS ($\mathrm{m/s^2}$)}} & 
\makecell[c]{\textbf{Longitudinal}\\\textbf{RMS ($\mathrm{m/s^2}$)}} \\
\midrule
CS      & 0.0 & \best{2.45} & 22.1 & \best{0.0} & 1.72 & \best{0.00} \\
RL-SC   & 0.0 & 4.63 & \second{55.4} & 12.5 & \second{1.47} & 1.79 \\
Opt-SP  & 0.0 & 4.74 & \second{55.4} & 12.5 & 1.46 & 1.79 \\
HAS     & \second{15.4} & 4.74 & 62.1 & 12.5 & 1.26 & 1.79 \\
\midrule
\textbf{Proposed} & \best{46.2} & \second{2.71} & \best{71.2} & \second{0.8} & \best{1.08} & \second{1.41} \\
\bottomrule
\end{tabularx}
\end{table*}

\textbf{Overall Performance (Table~\ref{tab:all}):} The proposed framework achieves the highest structural safety rate (71.2\%) and the highest avoidance rate (46.2\%) among all methods. While the CS baseline yields the shortest travel time, it also results in a very low structural safety rate (22.1\%). Compared with the longitudinal-only planners (RL-SC and Opt-SP), the proposed method reduces vertical RMS by approximately 26\% and lowers the collision rate from 12.5\% to 0.8\%, indicating a favorable overall trade-off between safety and comfort.

% =========================
% Table 2: Avoidable Scenarios
% =========================
\begin{table*}[!t]
\centering
\caption{Performance Statistics in 120 Avoidable Pothole Scenarios}
\label{tab:avoidable}
\renewcommand{\arraystretch}{1.2} % 保持行高一致
\footnotesize 
\setlength{\tabcolsep}{6pt} 
\begin{tabularx}{0.95\textwidth}{l | YYYYYY} 
\toprule
\textbf{Method} & 
\makecell[c]{\textbf{Avoidance}\\\textbf{Rate (\%)}} & 
\makecell[c]{\textbf{Travel}\\\textbf{Time (s)}} & 
\makecell[c]{\textbf{Structural Safety}\\\textbf{Rate (\%)}} & 
\makecell[c]{\textbf{Collision}\\\textbf{Rate (\%)}} & 
\makecell[c]{\textbf{Vertical}\\\textbf{RMS ($\mathrm{m/s^2}$)}} & 
\makecell[c]{\textbf{Longitudinal}\\\textbf{RMS ($\mathrm{m/s^2}$)}} \\
\midrule
CS      & 0.0 & \second{2.46} & 25.8 & \best{0.0} & 1.09 & \best{0.00} \\
RL-SC   & 0.0 & 4.75 & 56.7 & 11.7 & 1.13 & 1.81 \\
Opt-SP  & 0.0 & 4.75 & 56.7 & 11.7 & 1.13 & 1.81 \\
HAS     & \second{30.8} & 4.75 & \second{70.0} & 11.7 & \second{0.72} & 1.79 \\
\midrule
\textbf{Proposed} & \best{92.5} & \best{2.36} & \best{95.8} & \best{0.0} & \best{0.07} & \second{1.21} \\
\bottomrule
\end{tabularx}
\end{table*}

\textbf{Performance in Avoidable Scenarios (Table~\ref{tab:avoidable}):} The advantages of the hierarchical framework are most pronounced when lateral maneuvering is feasible. The proposed method achieves a 92.5\% avoidance rate, far exceeding the heuristic HAS baseline (30.8\%). By leveraging the nudging maneuvers visualized in Fig.~\ref{fig:avoidance_methods}, the vehicle virtually eliminates vertical impacts (Vert.\ RMS $\approx 0.07\,\mathrm{m/s^2}$) without the need for aggressive deceleration. Consequently, it reduces travel time by $\sim$50\% compared to methods that rely solely on braking, indicating that effective use of lateral maneuvering space is critical to improving both safety and efficiency.

% =========================
% Table 3: Unavoidable Scenarios
% =========================
\begin{table*}[!t]
\centering
\caption{Performance Statistics in 120 Unavoidable Pothole Scenarios}
\label{tab:unavoidable}
\renewcommand{\arraystretch}{1.2} % 保持行高一致
\footnotesize 
\setlength{\tabcolsep}{6pt} 
\begin{tabularx}{0.95\textwidth}{l | YYYYYY} 
\toprule
\textbf{Method} & 
\makecell[c]{\textbf{Avoidance}\\\textbf{Rate (\%)}} & 
\makecell[c]{\textbf{Travel}\\\textbf{Time (s)}} & 
\makecell[c]{\textbf{Structural Safety}\\\textbf{Rate (\%)}} & 
\makecell[c]{\textbf{Collision}\\\textbf{Rate (\%)}} & 
\makecell[c]{\textbf{Vertical}\\\textbf{RMS ($\mathrm{m/s^2}$)}} & 
\makecell[c]{\textbf{Longitudinal}\\\textbf{RMS ($\mathrm{m/s^2}$)}} \\
\midrule
CS      & 0.0 & \best{2.44} & 18.3 & \best{0.0} & 2.34 & \best{0.00} \\
RL-SC   & 0.0 & 4.51 & \best{54.2} & 13.3 & 1.81 & 1.77 \\
Opt-SP  & 0.0 & 4.73 & \best{54.2} & 13.3 & \best{1.79} & 1.78 \\
HAS     & 0.0 & 4.73 & \best{54.2} & 13.3 & \best{1.79} & 1.78 \\
\midrule
\textbf{Proposed} & 0.0 & \second{3.06} & \second{46.7} & \second{1.7} & \second{2.08} & \second{1.61} \\
\bottomrule
\end{tabularx}
\end{table*}

\textbf{Trade-offs in Unavoidable Scenarios (Table~\ref{tab:unavoidable}):} In scenarios where avoidance is geometrically impossible, a critical trade-off emerges between impact mitigation and traffic safety. The pure braking strategies (RL-SC, Opt-SP, HAS) achieve a slightly higher structural safety rate (54.2\%) by executing aggressive deceleration; however, this comes at the cost of a dangerously high rear-end collision rate (13.3\%). In contrast, the proposed method adopts a balanced strategy: it accepts a controlled deceleration (travel time 3.06~s) to mitigate damage (46.7\% structural safety) but avoids excessive braking that would endanger following traffic. This results in a collision rate of only 1.7\%, reflecting a learned safety hierarchy where preventing high-speed rear-end crashes takes precedence over maximizing pothole impact compliance.

\begin{figure*}[!t]
    \centering
    \subfloat[Vertical acceleration RMS.]{\includegraphics[width=0.32\linewidth]{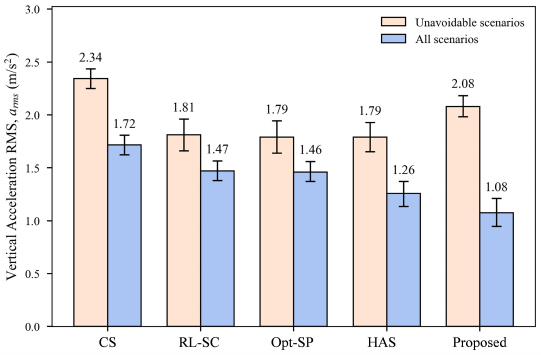}\label{fig:vert_acc}}
    \hfill
    \subfloat[Longitudinal acceleration RMS.]{\includegraphics[width=0.32\linewidth]{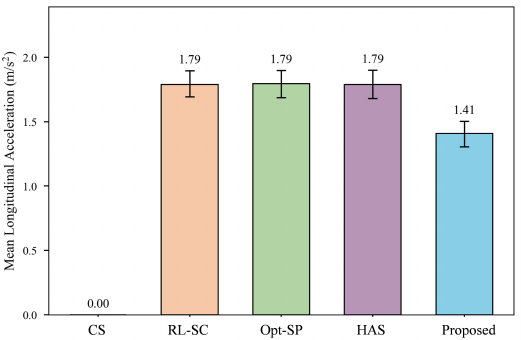}\label{fig:long_acc}}
    \hfill
    \subfloat[Impact force exceedance rate.]{\includegraphics[width=0.32\linewidth]{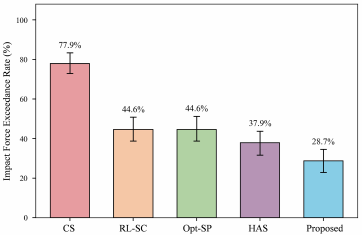}\label{fig:impact_rate}}
    
    \vspace{1em}
    
    \subfloat[Collision rate.]{\includegraphics[width=0.32\linewidth]{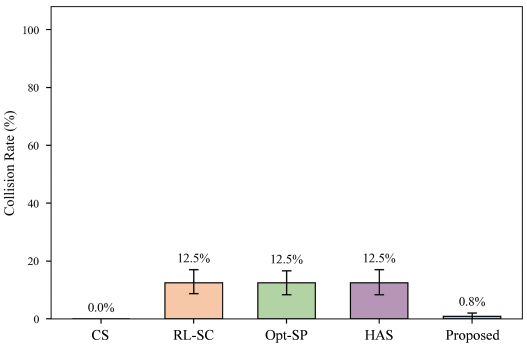}\label{fig:collision_rate}}
    \hfill
    \subfloat[Travel time.]{\includegraphics[width=0.32\linewidth]{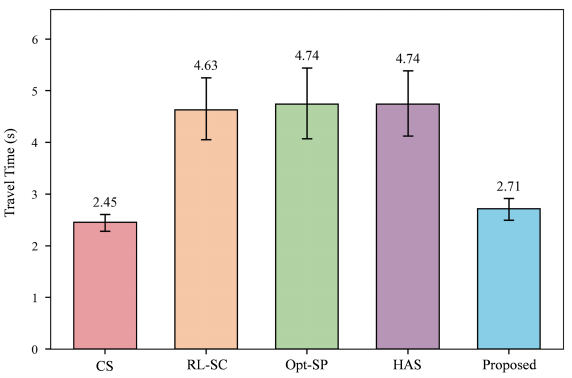}\label{fig:travel_time}}
    \hfill
    \subfloat[Avoidance rate.]{\includegraphics[width=0.32\linewidth]{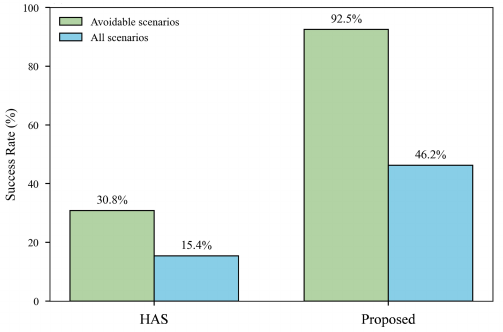}\label{fig:avoidance_rate}}
    \caption{Distributional comparison of key metrics across evaluation episodes. The proposed method exhibits a robust trade-off between impact safety, traffic safety, comfort, and efficiency.}
    \label{fig:metrics_comparison}
\end{figure*}

Finally, Fig.~\ref{fig:metrics_comparison} provides a holistic visualization of the performance landscape. As evident in Fig.~\ref{fig:metrics_comparison}(f), the proposed method (blue bar) shows the strongest overall performance in avoidance capability compared to the baselines. Furthermore, the collision rate comparison in Fig.~\ref{fig:metrics_comparison}(d) visually underscores the safety risks inherent in the pure-braking baselines, highlighting our framework's ability to maintain the most favorable balance between collision avoidance and structural safety across diverse operating conditions.

\subsection{Case Studies of Representative Scenarios}
\label{subsec:case_studies}

To complement the aggregate statistics in Section~\ref{subsec:comparative_analysis}, we present two representative episodes to interpret the behavioral mechanisms of the proposed method. The selected cases are drawn from the \emph{avoidable} and \emph{unavoidable} groups, respectively, and are chosen to highlight two recurrent decision patterns observed in evaluation: (i) lateral avoidance under elevated rear-end collision risk, and (ii) speed regulation to satisfy the impact-load limit when avoidance is infeasible. In both cases, we compare the proposed method against the baselines defined in Section~\ref{subsec:baselines} using the same scenario realization and traffic seed.

\subsubsection{Case I: Lateral avoidance under elevated rear-end collision risk}
\label{subsubsec:case1}

\begin{figure*}[!t]
    \centering
    \subfloat[]{\includegraphics[width=0.48\linewidth]{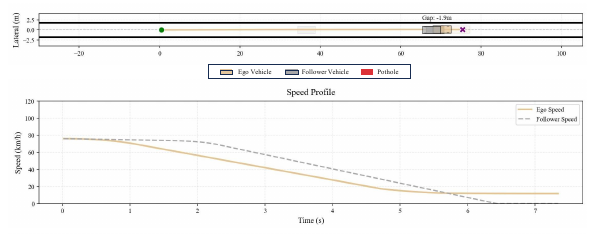}\label{fig:case1_has}}
    \hfill
    \subfloat[]{\includegraphics[width=0.48\linewidth]{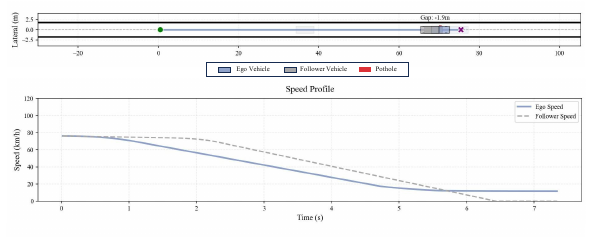}\label{fig:case1_optsp}}

    \vspace{0.6em}

    \subfloat[]{\includegraphics[width=0.48\linewidth]{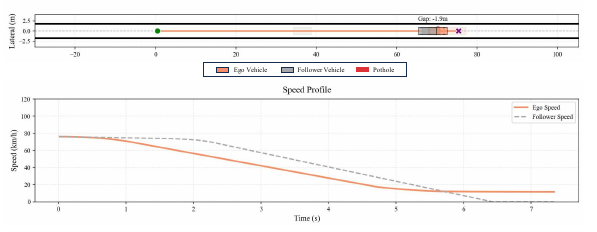}\label{fig:case1_rlsc}}
    \hfill
    \subfloat[]{\includegraphics[width=0.48\linewidth]{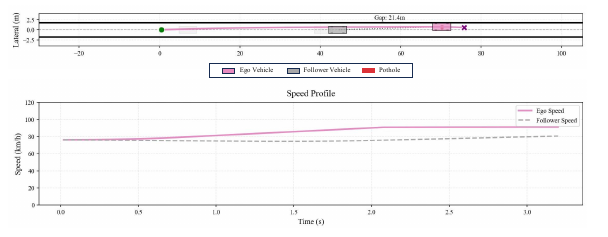}\label{fig:case1_prop}}

    \caption{Case I (\emph{avoidable}, $v_{\mathrm{init}}=76.1~\mathrm{km/h}$): Comparison of baseline strategies and the proposed method under tight car-following constraints. (a) HAS. (b) Opt-SP. (c) RL-SC. (d) Proposed method.}
    \label{fig:case_study_1}
\end{figure*}

This case belongs to the \emph{avoidable} group, where a deep pothole can be bypassed within lane boundaries. The ego vehicle approaches the hazard at $v_{\mathrm{init}}=76.1~\mathrm{km/h}$ with a closely following vehicle, creating a constrained car-following condition with limited braking margin. Fig.~\ref{fig:case_study_1} compares the time histories and outcomes of different strategies.

Across baselines, the dominant response is aggressive deceleration to reduce the pothole-induced vertical excitation. While this behavior mitigates impact severity, it does not explicitly account for rear-end collision risk under tight car-following conditions. Consequently, the rapid speed reduction leads to a violation of the rear-end safety margin and results in a rear-end collision in this episode.

In contrast, the proposed method adopts a joint decision that couples lateral feasibility with interaction safety. Rather than relying on strong braking, it executes a controlled lateral bypass maneuver while maintaining a flow-consistent speed profile. This behavior preserves the rear-end safety margin and simultaneously avoids the pothole, yielding zero structural impact in this episode. This case illustrates that, when avoidance is feasible, joint lateral-longitudinal planning can preserve rear-end safety while eliminating the impact load.

\subsubsection{Case II: Impact-load-constrained speed regulation in unavoidable hazards}
\label{subsubsec:case2}

\begin{figure*}[!t]
    \centering
    \subfloat[]{\includegraphics[width=0.48\linewidth]{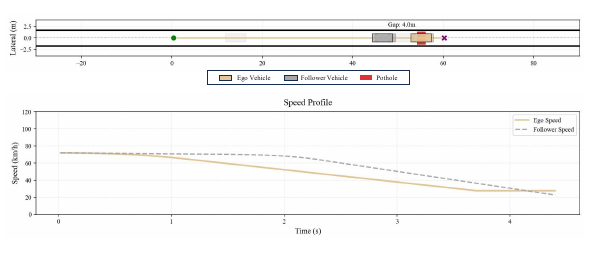}\label{fig:case2_has}}
    \hfill
    \subfloat[]{\includegraphics[width=0.48\linewidth]{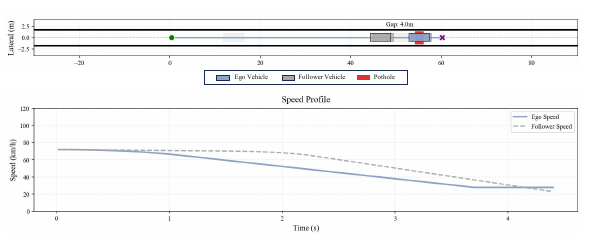}\label{fig:case2_optsp}}

    \vspace{0.6em}

    \subfloat[]{\includegraphics[width=0.48\linewidth]{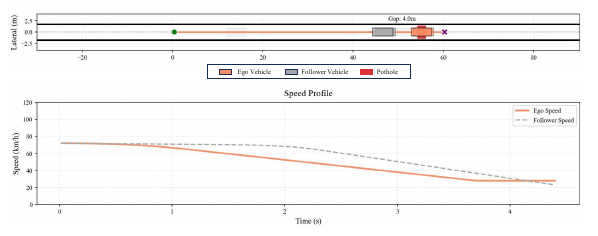}\label{fig:case2_rlsc}}
    \hfill
    \subfloat[]{\includegraphics[width=0.48\linewidth]{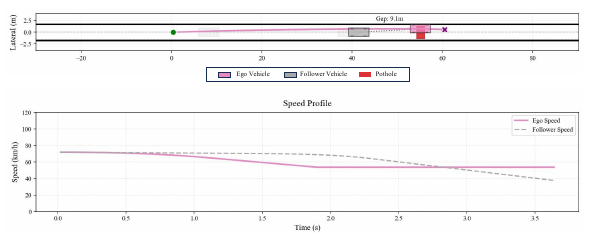}\label{fig:case2_prop}}

    \caption{Case II (\emph{unavoidable}, $v_{\mathrm{init}}=68.1~\mathrm{km/h}$): Comparison under an unavoidable hazard, highlighting impact-load-constrained speed regulation. (a) HAS. (b) Opt-SP. (c) RL-SC. (d) Proposed method.}
    \label{fig:case_study_2}
\end{figure*}

This case belongs to the \emph{unavoidable} group, where the hazard spans the lane and geometric avoidance is infeasible. The ego vehicle approaches an unavoidable pothole at $v_{\mathrm{init}}=68.1~\mathrm{km/h}$. Under this setting, the decision primarily concerns how to reduce the impact load without inducing excessive rear-end risk. Fig.~\ref{fig:case_study_2} compares the resulting speed profiles and safety outcomes.

Baseline methods again reduce speed to a near-minimum level to minimize impact. Although such over-braking can reduce the instantaneous impact load, it substantially compresses the rear gap and increases rear-end risk, which is undesirable in mixed traffic.

The proposed method employs an impact-load-constrained regulation strategy, guided by the calibrated structural load limit ($F_{\mathrm{lim}}=15.15~\mathrm{kN}$). It decelerates to a compliant yet non-extreme speed of $29.1~\mathrm{km/h}$, resulting in an impact force of $12.53~\mathrm{kN}$ (below $F_{\mathrm{lim}}$). By avoiding excessive braking, the proposed method preserves a larger rear-end safety margin while ensuring compliance with structural safety constraints. This case further illustrates the trade-off discussed in Section~\ref{subsec:comparative_analysis}: in unavoidable hazards, strict impact minimization may conflict with interaction safety. The proposed approach addresses this trade-off through risk-prioritized, threshold-aware speed regulation.

\section{Discussions}
\label{sec:discussion}

The experimental results in Section~\ref{sec:results} show that the proposed framework consistently outperforms the baseline strategies in single-hazard scenarios. In this section, we further interpret these findings by examining policy generalization in sequential hazard scenarios.

\subsection{Generalization to Sequential Hazards}
\label{subsec:generalization}

A critical challenge for learning-based control systems is overfitting to the training scenarios. While the curriculum in this study primarily focused on single-pothole cases, real-world road surfaces often contain clustered or consecutive defects. To assess policy generalization, we evaluated the agent in a sequential hazard scenario with multiple potholes of varying geometries.

%Fig.~\ref{fig:multi_pothole} visualizes the vehicle's trajectory and velocity profile in this unseen complex environment. As illustrated, the agent successfully executes a continuous chain of tactical maneuvers: it performs a lateral nudge to bypass the first hazard, stabilizes its yaw motion, and immediately prepares for the subsequent avoidance.

\begin{figure}[!t]
    \centering
    \includegraphics[width=\columnwidth]{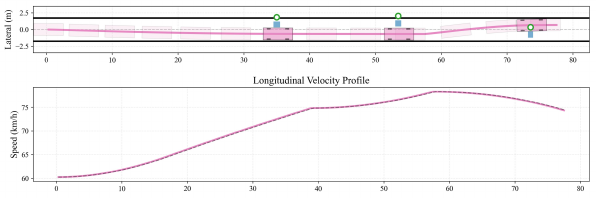}
    \caption{Generalization test in a sequential hazard scenario. The agent navigates multiple potholes in sequence and accelerates during safe intervals, indicating policy generalization beyond the single-hazard training setting.}
    \label{fig:multi_pothole}
\end{figure}

As shown in Fig.~\ref{fig:multi_pothole}, the agent successfully executes a sequence of tactical maneuvers, including a lateral nudge to bypass the first hazard, yaw stabilization, and immediate preparation for the next avoidance. The velocity profile highlights an active ``efficiency-recovery'' strategy, where the agent avoids a passive ``stop-and-go'' approach by accelerating during safe intervals between potholes to restore traffic flow speed. 

This behavior suggests that the learned policy generalizes beyond the specific training scenarios, rather than simply reproducing memorized responses. By adapting its speed and lateral behavior across successive hazards, the agent shows promising robustness in the presence of clustered road-surface defects.

%Most notably, the velocity profile reveals an active ``efficiency-recovery'' behavior. The agent does not adopt a passive ``stop-and-go'' strategy; instead, it actively accelerates during the safe intervals between potholes to restore traffic flow speed. This behavior suggests that the agent has not merely memorized specific state-action pairs but has internalized a generalized \emph{dynamic driving skill}. It dynamically assesses the risk horizon and seizes opportunities to improve efficiency, demonstrating robust adaptability to unstructured infrastructure deterioration.

\subsection{Limitations}
\label{subsec:limitations}

%While the proposed framework demonstrates efficacy in mitigating terrain-induced disturbances, several modeling abstractions warrant further investigation. First, the current formulation is predicated on \textit{deterministic perception inputs}, assuming precise knowledge of pothole geometry. In operational scenarios, however, depth estimation is inherently subject to stochastic sensor noise and occlusion. Consequently, future iterations should integrate uncertainty-aware planning paradigms (e.g., belief-space planning) to robustly handle perceptual ambiguity. Second, although the quarter-car model captures vertical dynamics efficiently, it inherently decouples vertical motion from lateral and roll dynamics. To bridge the simulation-to-reality gap, validating the algorithm using high-fidelity multi-body dynamics simulations or scaled hardware testbeds remains a critical step toward deployment in complex dynamic environments.

While our proposed framework effectively mitigates pothole-induced disturbances, certain modeling abstractions require further investigation. First, the current formulation relies on  \textit{deterministic perception inputs}, assuming precise knowledge of pothole geometry. In practice, depth estimation is subject to stochastic sensor noise and occlusion, necessitating the integration of uncertainty-aware planning paradigms (e.g., belief-space planning) in future work to address perceptual ambiguity. Second, although the quarter-car model effectively captures vertical dynamics, it decouples vertical motion from lateral and roll dynamics. To bridge the simulation-to-reality gap, validating the algorithm through high-fidelity multi-body dynamics simulations or scaled hardware testbeds remains critical for deployment in complex dynamic environments.

\section{Conclusions}
\label{sec:conclusion}

This paper presented a vehicle-dynamics-oriented hierarchical decision-making framework for pothole-aware autonomous driving. In contrast to existing methods that mainly emphasize geometric avoidance or isolated suspension control, the proposed approach combines a mechanism-informed tire--pothole impact model with curriculum-based reinforcement learning, enabling the agent to balance structural safety, ride comfort, and traffic safety under tight physical constraints.

The primary findings from this study are:

\begin{enumerate}
\item \textbf{Improved structural safety:} In unavoidable hazard scenarios, the proposed framework achieves a structural safety rate of 46.7\%, indicating a clear improvement over the passive constant-speed baseline (22.1\%). This result shows that the learned policy can effectively reduce excessive impact loads even when full geometric avoidance is not possible.

\item \textbf{Improved interaction safety:} In unavoidable scenarios, the proposed method reduces the rear-end collision rate to 1.7\%, compared with 13.3\% for the braking-heavy baselines. This result indicates that the learned policy avoids overly aggressive braking and instead balances impact mitigation with interaction safety in surrounding traffic.

\item \textbf{Emergent adaptive behavior:} The learned policy exhibits diverse behaviors, including lateral nudging, straddling, and threshold-aware speed regulation. It also generalizes to sequential hazard scenarios without additional retraining, adjusting its speed between successive potholes to recover traffic efficiency.

\end{enumerate}

In conclusion, this study bridges the gap between conventional geometric path planning and vehicle-road interaction mechanics, presenting a resilient solution for autonomous mobility in unpredictable infrastructure settings. This work shifts the safety paradigm from static obstacle avoidance to dynamic impact management. Future research will concentrate on integrating perceptual uncertainties (such as noisy depth estimation) and validating proposed algorithms on full-scale vehicle platforms to tackle real-world sensing and actuation latency issues.

% ==========================================
% BIBLIOGRAPHY
% ==========================================
\bibliographystyle{IEEEtranS}
\bibliography{references}

% ==========================================
% AUTHOR BIOGRAPHIES (IEEE Compliant Version)
% ==========================================

% 1. Xiang Wang
\begin{IEEEbiography}[{\includegraphics[width=1in,height=1.25in,clip,keepaspectratio]{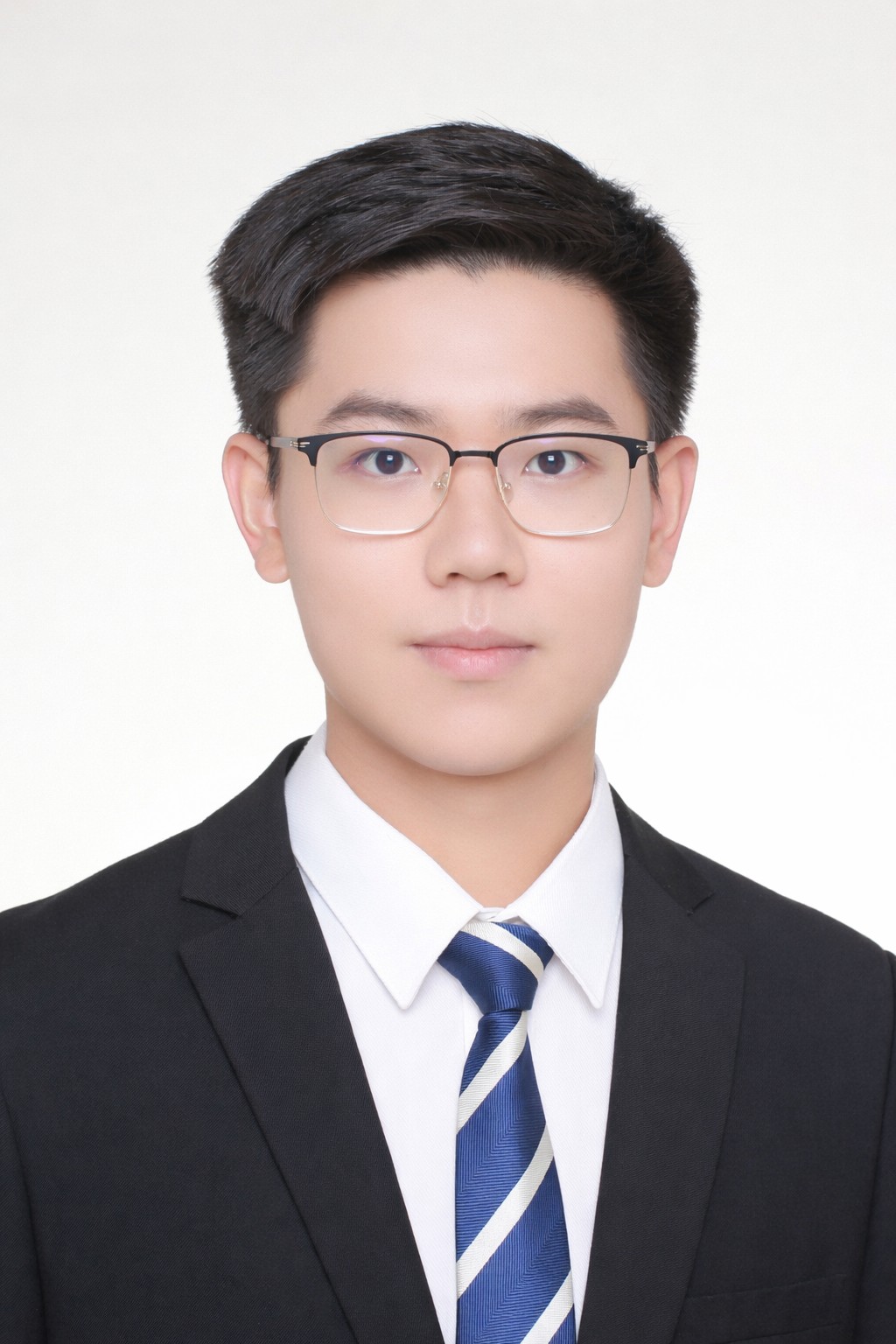}}]{Xiang Wang}
received the B.Sc. degree in transportation engineering from the Harbin Institute of Technology, Weihai, China, in 2024. He is currently working toward the M.Sc. degree with the College of Transportation Engineering, Tongji University, Shanghai, China. His research interests include motion planning, vehicle dynamics control, autonomous vehicle safety and intelligent vehicles.
\end{IEEEbiography}

% 2. Scott Piersall
\begin{IEEEbiography}[{\includegraphics[width=1in,height=1.25in,clip,keepaspectratio]{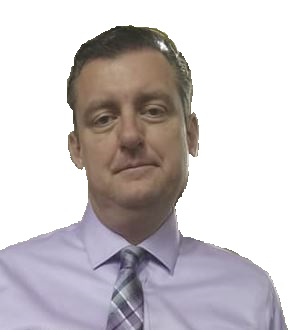}}]{Scott Piersall}
received the M.S. degree in computer science from the University of North Florida, Jacksonville, FL, USA, in 2000. He is currently working toward the Ph.D. degree in computer science at the University of Central Florida, Orlando, FL, USA. 

He is a Visiting Instructor with the School of Computing, University of North Florida, where he teaches undergraduate courses in systems programming, enterprise COBOL, and graduate courses in parallel computing. His research interests include machine learning, large language models, program analysis, parallel computing, and distributed systems.
\end{IEEEbiography}

% 3. Zihang Zou
\begin{IEEEbiography}[{\includegraphics[width=1in,height=1.25in,clip,keepaspectratio]{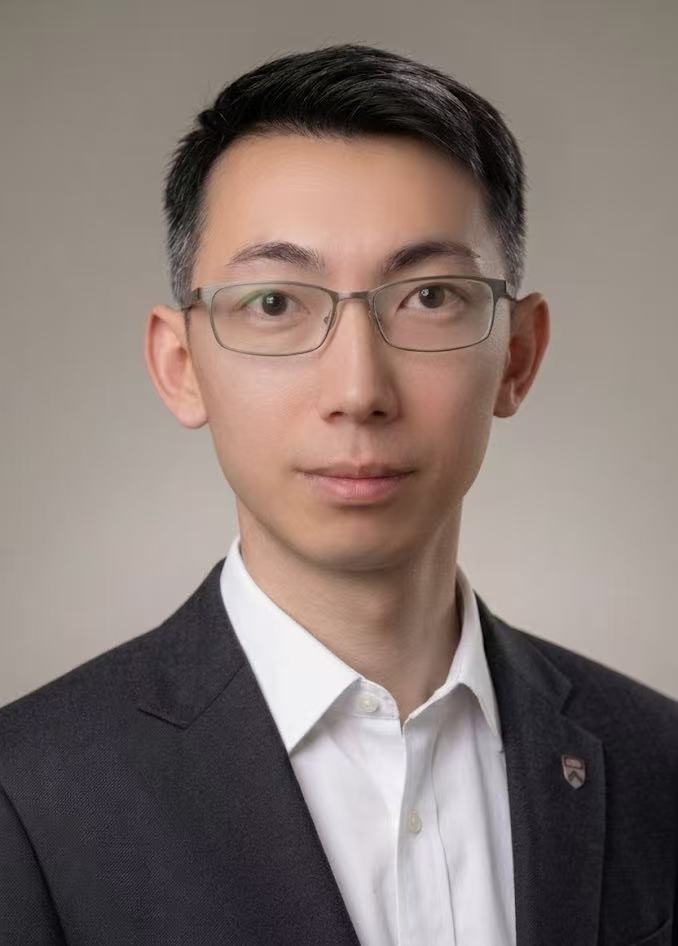}}]{Zihang Zou}
received the B.Sc. degree in electrical and electronics engineering from Sun Yat-sen University, Guangzhou, China, in 2010, the M.Sc. degree in electrical and electronics engineering from the University of Central Florida (UCF), Orlando, FL, USA, in 2011, and the Ph.D. degree from the College of Engineering and Computer Science, UCF, in 2025. 

He is currently a Lead Research Scientist at Optixway AI, USA. His research focuses on safe, reliable, and legally compliant artificial intelligence, with interests spanning trustworthy generative AI, intellectual property protection for generative models, lawful autonomous driving, transportation safety, and AI systems designed for real-world deployment. His work covers neural watermarking and ownership verification for generative models, crash-risk prediction, and traffic-law-aware AI, with a broader emphasis on aligning AI systems with real-world rules, operational constraints, and societal needs. More broadly, he is interested in advancing robust, compliant, and socially responsible AI for safety-critical and infrastructure-related applications.
\end{IEEEbiography}

% 4. Liqiang Wang (修正了高度并加入微调)
% 注意：如果图片还是低，可以把 \vspace{-1mm} 改成 -2mm
\begin{IEEEbiography}[{\vspace{-1mm}\includegraphics[width=1in,height=1.25in,clip,keepaspectratio]{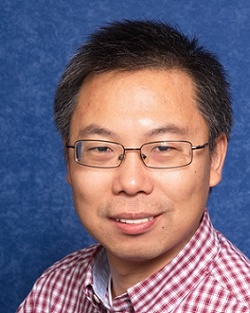}}]{Liqiang Wang}
received the Ph.D. degree in computer science from Stony Brook University, Stony Brook, NY, USA, in 2006. He was a Faculty Member with the Department of Computer Science, University of Wyoming, Laramie, WY, USA, from 2006 to 2015, and a Visiting Research Scientist with the IBM T.J. Watson Research Center from 2012 to 2013. 

He is currently a Professor with the Department of Computer Science, University of Central Florida, Orlando, FL, USA, where he is also the Director of the Big Data Lab. His research focuses on big data computing and analytics, including improving model accuracy and security, and optimizing performance of parallel computing systems. Dr. Wang was a recipient of the NSF CAREER Award in 2011, the Castagne Faculty Fellowship (2013--2015), and the UCF Mid-Career Refresh Award in 2020.
\end{IEEEbiography}

% 5. Rongjie Yu
\begin{IEEEbiography}[{\includegraphics[width=1in,height=1.25in,clip,keepaspectratio]{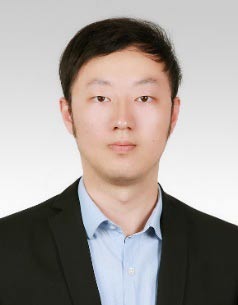}}]{Rongjie Yu}
received the B.Sc. degree in civil engineering from Tongji University, Shanghai, China, in 2010, and the M.Sc. degree in transportation engineering and the Ph.D. degree from the University of Central Florida, Orlando, FL, USA, in 2012 and 2013, respectively. 

He is currently a Professor with the College of Transportation Engineering, Tongji University, Shanghai, China. His research interests include traffic safety analysis, driving behavior modeling, proactive safety management, and autonomous vehicle safety evaluation.
\end{IEEEbiography}

\end{document}